\documentclass[journal]{IEEEtran}

\usepackage{comment}
\usepackage{graphicx}
\usepackage{cite}
\usepackage{picinpar}
\usepackage{amsmath,amsfonts}
\usepackage{url}
\usepackage{flushend}
\usepackage{colortbl}
\usepackage{soul}
\usepackage{multirow}
\usepackage{pifont}
\usepackage{color}
\usepackage{alltt}
\usepackage[hidelinks]{hyperref}
\usepackage{enumerate}
\usepackage{siunitx}
\usepackage{breakurl}
\usepackage{epstopdf}
\usepackage{pbox}
\usepackage{xcolor}
\usepackage[utf8]{inputenc}
\usepackage[english]{babel}
\usepackage[linesnumbered,ruled,vlined]{algorithm2e}
\usepackage{booktabs}
\usepackage{makecell}
\usepackage{float}
\usepackage{dblfloatfix}
\usepackage{threeparttable}
\usepackage{hyperref}
\hypersetup{
    colorlinks=true,
    linkcolor=orange,
    citecolor=blue,
    filecolor=magenta,      
    urlcolor=blue,
}

\usepackage[figurename=Fig.]{caption}
\SetKwInput{KwInput}{Input}
\SetKwInput{KwOutput}{Output}

\title{\fontsize{20pt}{22pt}\selectfont SafeDrive: Knowledge- and Data-Driven Risk-Sensitive\\ 
Decision-Making for Autonomous Vehicles with Large Language Models}

\author{
    \fontsize{12pt}{12pt}\selectfont
    Zhiyuan Zhou, Heye Huang*, Boqi Li, Shiyue Zhao, Yao Mu, Jianqiang Wang%
    \vspace{0.3em} 

    \thanks{\fontsize{8.5pt}{10pt}\selectfont
    Z. Zhou is with the Department of Financial Engineering, University of Southern California, CA 90089, USA.  
    H. Huang is with the Department of Civil and Environmental Engineering, University of Wisconsin-Madison, WI 53706, USA.  
    B. Li is with the Department of Civil and Environmental Engineering, University of Michigan, Ann Arbor, MI 48105, USA.  
    S. Zhao and J. Wang are with the School of Vehicle and Mobility, Tsinghua University, Beijing, 100084, China. 
    Y. Mu is with the Department of Computer Science, The University of Hong Kong, Hong Kong, China.  
    *Corresponding author: H. Huang (e-mail: hhuang468@wisc.edu).}
}

\begin{document}

\maketitle

\begin{abstract}
\noindent\fontsize{9pt}{11pt}\selectfont
Recent advancements in autonomous vehicles (AVs) leverage Large Language Models (LLMs) to perform well in normal driving scenarios. However, ensuring safety in dynamic, high-risk environments and managing safety-critical long-tail events remains a significant challenge. To address these issues, we propose SafeDrive, a knowledge- and data-driven risk-sensitive decision-making framework, to enhance AV safety and adaptability. The proposed framework introduces a modular system comprising: (1) a Risk Module for comprehensive quantification of multi-factor coupled risks involving driver, vehicle, and road interactions; (2) a Memory Module for storing and retrieving typical scenarios to improve adaptability; (3) a LLM-powered Reasoning Module for context-aware safety decision-making; and (4) a Reflection Module for refining decisions through iterative learning. By integrating knowledge-driven insights with adaptive learning mechanisms, the framework ensures robust decision-making under uncertain conditions. Extensive evaluations on real-world traffic datasets characterized by dynamic and high-risk scenarios, including highways (HighD), intersections (InD), and roundabouts (RounD), validate the framework’s ability to enhance decision-making safety (achieving a 100\% safety rate), replicate human-like driving behaviors (with decision alignment exceeding 85\%), and adapt effectively to unpredictable scenarios. The proposed framework of SafeDrive establishes a novel paradigm for integrating knowledge- and data-driven methods, highlighting significant potential to improve the safety and adaptability of autonomous driving in long-tail or high-risk traffic scenarios. Project page: \url{https://mezzi33.github.io/SafeDrive/}.
\end{abstract}


{}

\section{Introduction}
\IEEEPARstart{A}{utonomous} vehicles (AVs) have advanced significantly in recent years, achieving the ability to operate safely in most traffic scenarios \cite{wang2021towards}. To enhance their capacity to handle diverse scenarios and progress toward higher levels of automation, data-driven paradigms leverage vast real-world driving data and learning algorithms, achieving remarkable accuracy in tasks such as prediction and decision-making \cite{o2022neural, elallid2022comprehensive}. However, data-driven paradigms face several limitations: 1) data bias, which prioritizes common scenarios while neglecting rare corner cases; and 2) a lack of interpretability, with AVs operating as black boxes, making causal relationships unclear and safety difficult to guarantee. These limitations hinder AVs' ability to make rational, human-acceptable decisions in rare and unpredictable long-tail scenarios \cite{ding2024adaptive, he2024agile}.

To achieve fundamental progress in autonomous driving, enabling AVs to acquire human-like interactive driving abilities through understanding and learning from human behavior is crucial \cite{fu2024drive}. With their human-like experience and common sense, large language models (LLMs) as decision-making agents present a promising direction \cite{cui2024receive}. By incorporating the human-like capabilities of learning and reasoning into the system, LLM-based agents aim to enhance the contextual understanding and adaptability of AVs.

Despite the promising capabilities of LLM-powered agents, they can occasionally display overconfidence, leading to risky actions \cite{Wang_2024}. While the safety performance of LLM-driven agents has been studied extensively, most research focuses on simple simulated highway environments, where risk sources are limited, and interactions are minimal. However, real-world traffic is complex, dynamic, and coupled, with frequent potential conflicts, making it more challenging for LLM-based agents to accurately identify all potential risks and ensure safety \cite{fu2024drive}. Consequently, the following research questions are crucial for advancing higher-level LLM-based driving agents:
\begin{enumerate}
    \item \textbf{RQ1. How can we effectively model and quantify the coupled risks in safety-critical environments?}
    
    \item \textbf{RQ2. How can we guide the LLM-based agents to derive safe and human-like driving behavior?}
\end{enumerate}

\begin{figure*}[h]
\centerline{\includegraphics[width=1\textwidth]{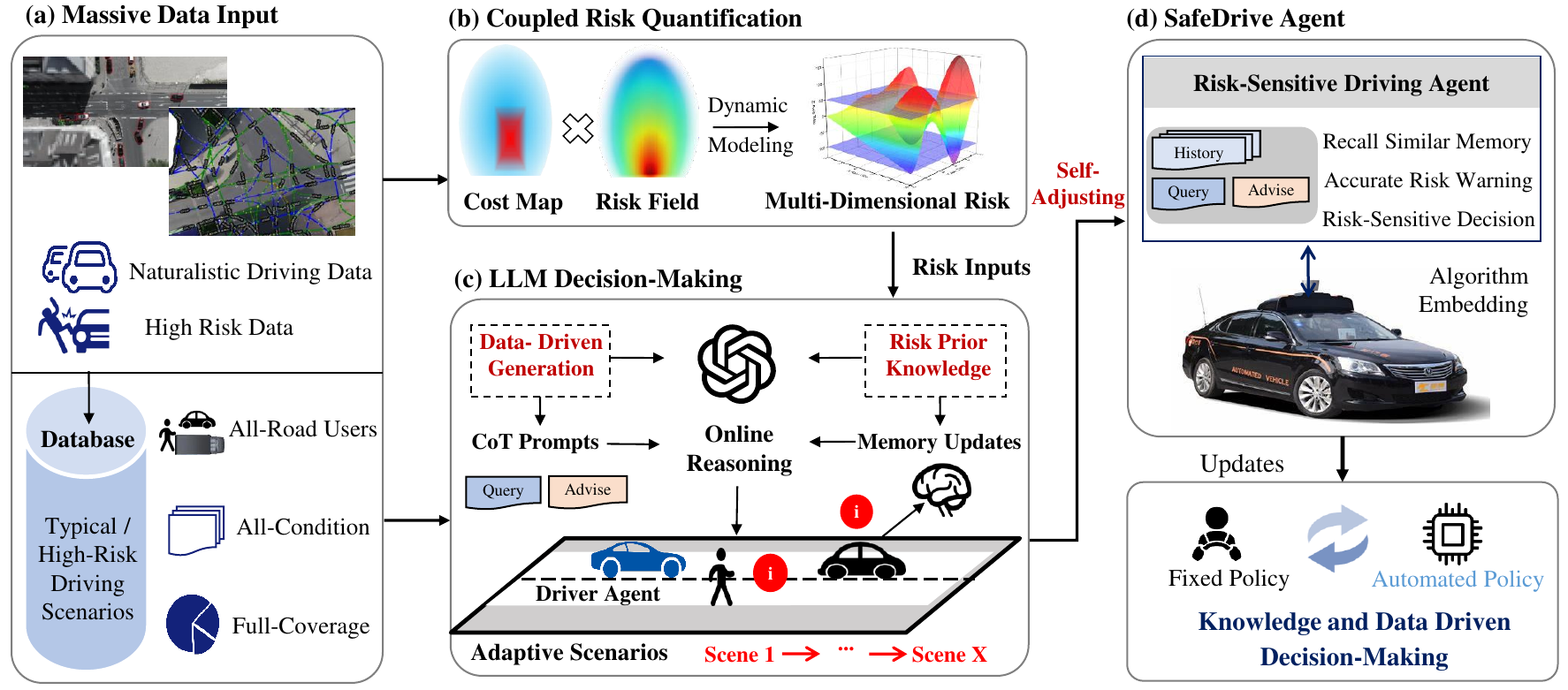}}
    \caption{Overview of SafeDrive: the knowledge- and data-driven risk-sensitive decision-making framework. 
    \textbf{Input:} Hand-labeled high-risk scenario descriptions from real-world datasets. \textbf{Process:} The coupled risk quantification model generates multi-dimensional risk assessments based on real-time trajectory data. The data-driven scenario descriptions, combined with risk prior knowledge and past experiences, are processed by the LLM-based agent using chain-of-thought (CoT) reasoning for adaptive decision-making. \textbf{Output:} An action decision (accelerate, decelerate, change lanes, or idle) generated by the LLM agent, checked through a reflection module for correction, and stored in the memory database for future retrieval, enabling continuous improvement.
    }
    \label{fig_framework}
\end{figure*}

To address the two key questions, this paper proposes a novel knowledge- and data-driven risk-sensitive decision-making framework: SafeDrive. It introduces a comprehensive risk quantification model that considers multiple risk sources from the human, vehicle, and road, providing precise and universally applicable risk assessments. Additionally, it leverages real-world driving data and scenario-risk knowledge to enable LLM-based safe decision-making. The SafeDrive Framework operates through a four-module loop, with OpenAI ChatGPT-4 (GPT-4) as the driving agent. The modules are as follows: 
\begin{enumerate}
    \item \textbf{Risk Module:} outputs comprehensive and accurate risk assessments for various scenarios.
    \item \textbf{Memory Module:} stores and retrieves relevant past experiences according to the scene description and risk levels output by the Risk Module. 
    \item \textbf{Reasoning Module:} combines prior risk knowledge with dynamic contextual learning for AVs' reasoning.
    \item \textbf{Reflection Module}: assesses and corrects any incorrect decisions, facilitating a deeper reflection process. 
\end{enumerate}

These modules form a robust, continuous learning system ensuring safe navigation in complex traffic. The SafeDrive framework's effectiveness is validated through extensive experiments on diverse real-world datasets, including HighD (highway) \cite{highd}, InD (intersection) \cite{ind}, and RounD (roundabout) \cite{round}. In summary, our contributions are as follows:
\begin{enumerate}
    \item We propose a unified risk quantification model that provides comprehensive and omnidirectional assessment of multi-factor coupled risks, enabling real-time, continuous driving risk quantification.
    
    \item We introduce a knowledge- and data-driven, risk-sensitive decision-making framework that combines risk assessments with dynamic contextual learning, enhancing AV safety and interpretability in uncertain scenarios.
    
    \item Our framework achieves a 100\% safety rate and over 85\% alignment with human decision-making, demonstrating strong adaptability in  high-risk scenarios while ensuring long-term reliability across diverse traffic conditions.
\end{enumerate}

\section{Related Works}
\textbf{Risk Quantification.} Risk quantification is essential for collision prevention in AVs. Classical methods considering vehicle dynamics such as Time to Collision (TTC)\cite{katrakazas2019new}, Time Headway (THW)\cite{vogel2003comparison}, Time to React (TTR)\cite{wagner2018using}, and Lane-Crossing Time (TLC)\cite{lin1996time} are widely applied in traffic scenarios due to their simplicity and ease of implementation. However, these methods often fall short in dynamic, multi-dimensional environments, where risk factors change rapidly and interact in complex ways \cite{vogel2003comparison}. To address these limitations, Shalev-Shwartz et al. propose responsibility-sensitive safety (RSS) ~\cite{shalev2017formal}, a model designed for more interpretable, white-box safety assurance. However, dilemmas still exist, such as the determination of a large number of parameters.

To overcome these limitations, advanced methods have been proposed. The Artificial Potential Field (APF) method~\cite{warren1989global} uses potential fields to model vehicle risks, enabling basic collision avoidance. Gerdes et al.\cite{gerdes2001unified} extended the APF by incorporating lane markings to create detailed risk maps. Wang et al.\cite{huang2020probabilistic, wang2020driving} integrate road potential fields with vehicle dynamics and driver behavior, improving the accuracy of risk simulations and reducing collision risks in complex scenarios.  However, these approaches often focus on current traffic states, rely on numerous parameters, and lack adaptability in uncertain environments. Kolekar et al.\cite{kolekar2020human} introduce the Driver’s Risk Field (DRF), a two-dimensional model that incorporates drivers' subjective risk perceptions based on probabilistic beliefs. By integrating subjective risk assessments, these improved APF and DRF methods better simulate traffic system dynamics and enhance multi-dimensional risk evaluation. However, the DRF proposed by Kolekar et al.\cite{kolekar2021risk} only accounts for the risk in the heading direction, the forward-facing half circle of the vehicle, falling short in providing a comprehensive omnidirectional risk quantification. This limitation can be compensated by the risk quantification model introduced in this paper.

\textbf{LLM in Decision Making.} Decision-making is crucial for autonomous driving, as it directly determines the vehicle's ability to navigate complex, dynamic, and high-conflict traffic scenarios safely and efficiently \cite{huang2024general}. Traditional data-driven decision-making methods have inherent limitations. These algorithms are often regarded as black boxes, and their sensitivity to data bias, difficulty in handling long-tail scenarios, and lack of interpretability present significant challenges in providing human-understandable explanations for their decisions, especially when adapting to data-scarce long-tail scenarios\cite{wu2024adaptive,zhang2024fuzzy}. 

Advancements in LLMs offer valuable insights for addressing decision-making challenges in autonomous driving. LLMs demonstrate human-level capabilities in perception, prediction, and planning\cite{mu2024embodiedgpt}. LLMs, when coupled with a vector database as memory, showcase impressive enhancements in analytical capabilities in certain areas\cite{jing2024largelanguagemodelsmeet}. Li et al. propose the concept of knowledge-driven autonomous driving, showing that LLMs can enhance real-world decision-making through common-sense knowledge and driving experience \cite{li2023towards}. Wen et al. proposed the DiLu framework, which integrates reasoning and reflection for knowledge-driven, continuously evolving decision-making, outperforming reinforcement learning methods \cite{wen2023dilu}. Jiang et al. employ DiLu as the foundation and develop a knowledge-driven multi-agent framework for autonomous driving, demonstrating its efficiency and accuracy across various driving tasks \cite{jiang2024koma}. Fang et al.~\cite{fang2024towards} focus on using LLMs as agents for cooperative driving in different scenarios. 
Recent advancements also highlight LLMs’ potential for multimodal reasoning. Hwang et al. \cite{hwang2024emma} introduce EMMA, an End-to-End Multimodal Model using pretrained LLMs for motion planning, achieving state-of-the-art results with nuScenes and WOMD. However, its reliance on image inputs and high computational costs pose challenges. Sinha et al. \cite{sinha2024real} propose a two-stage framework combining a fast anomaly classifier with fallback reasoning for real-time anomaly detection and reactive planning, demonstrating robustness in simulations. These studies emphasize the potential of LLMs in decision-making of AVs, where real-time reasoning and adaptability are essential. However, most research focuses on simple scenarios and lacks adaptability in high-conflict environments. This paper explores LLMs’ knowledge-driven capabilities to address decision-making challenges in dynamic and highly-uncertain scenarios, filling a critical gap in the field.

\section{Framework}
In this section, we propose SafeDrive, a knowledge- and data-driven risk-sensitive decision-making framework based on LLM, as shown in \autoref{fig_framework}. SafeDrive combines naturalistic driving data and high-risk scenarios to enable AVs to make adaptive, safe decisions in complex, dynamic environments.

The framework begins with massive data input (Figure 1a), combining all-road users, all-condition scenarios, and full-coverage data into a comprehensive database of typical and high-risk driving scenarios. In the coupled risk quantification module (Fig. 1b), advanced risk modeling, including cost maps and multi-dimensional risk fields, dynamically quantifies risks, providing detailed inputs for decision-making. The LLM decision-making module (Fig. 1c) uses data-driven generation, risk prior knowledge, and chain-of-thought (CoT) reasoning to generate real-time, risk-sensitive decisions. Additionally, adaptive memory updates ensure that similar experiences can be recalled to refine the decision process. These decisions are embedded into a risk-sensitive driving agent (Fig. 1d) that delivers accurate risk warnings, recalls past experiences, and makes adaptive decisions. The self-adjusting system ensures real-time risk identification and enables continuous updates to driving policies through a closed-loop reflection mechanism.

Overall, SafeDrive enhances real-time responsiveness, decision safety, and adaptability, addressing challenges in high-risk, unpredictable scenarios.

\section{Methodology}
In this section, we aim to address the two crucial research questions for achieving higher-level AVs. First, we introduce a coupled risk quantification model. Second, we propose a risk-sensitive decision-making method with the integration of LLM and the risk model.
\subsection{Coupled Risk Quantification}
\textbf{RQ1.  How can we effectively model and quantify the coupled risks in safety-critical environments?}

The concept of perceived risk, as defined by Naatanen and Summala et al.~\cite{naatanen1976road}, is the product of the subjective probability that an event will occur and the consequence of that event. In this paper, we employ a dynamic Driver Risk Field (DRF) model that adapts to vehicle speed and steering dynamics, inspired by Kolekar et al. The DRF represents the driver’s subjective belief about future positions, assigning higher risk near the ego vehicle and decreasing with distance. Event consequences are quantified by assigning experimentally determined costs to objects in the scene based on their danger level, independent of subjective assessments. The overall Quantified Perceived Risk (QPR) is computed as the sum of the event costs and the DRF across all grid points. This approach effectively captures uncertainties in driver perception and actions, offering a comprehensive measure of driving risk.

\begin{figure}[ht]
\centering
\includegraphics[width=1\linewidth]{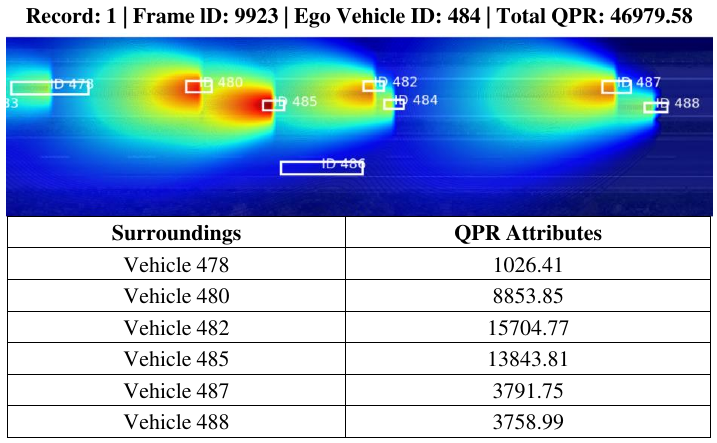}
\caption{The omnidirectional risk quantification of surrounding vehicles: QPR attributes include the risk cost of each vehicle that has a mutual impact with the ego vehicle, based on the DRF distribution.}
\label{fig_qpr_attributes}
\end{figure}

\textbf{Driver Risk Field.} 
This work extends the DRF to account for dynamic changes based on vehicle speed and steering angle. The DRF is computed using a kinematic car model, where the predicted path depends on the vehicle's position \((x_{\text{car}}, y_{\text{car}})\), heading \(\phi_{\text{car}}\), and steering angle \(\delta\). The radius of the predicted travel arc, assuming a constant steering angle, is given by:
\begin{equation}
R_{\text{car}} = \frac{L}{\tan(\delta)}
\end{equation}
where \(L\) is the car's wheelbase. Using the vehicle's position and arc radius, the center of the turning circle \((x_c, y_c)\) is found, which is used to compute the arc length \(s\), representing the distance along the path.

The DRF is modeled as a torus with a Gaussian cross-section:
\begin{equation}
z(x,y) = a \exp\left(-\frac{((x-x_c)^2 + (y-y_c)^2 - R_{\text{car}}^2)^2}{2\sigma^2}\right)
\end{equation}
where \(a\) is the height and \(\sigma\) the width of the Gaussian, both functions of the arc length \(s\). The height \(a(s)\) is a parabolic function of \(s\), given by:
\begin{equation}
a(s) = p - v t_{\text{la}}
\end{equation}
where \(t_{\text{la}}\) scales with vehicle speed \(v\) and \(p\) determines the parabola’s steepness. The width \(\sigma_i\) is a linear function of arc length \(s\), given by:
\begin{equation}
\sigma_i = (m + k_i |\delta|)s + c, \quad
i = \begin{cases} 
1 & \text{(inner } \sigma\text{)} \\
2 & \text{(outer } \sigma\text{)}
\end{cases}
\end{equation}
Here, \(c\) defines the DRF’s width at the vehicle’s location, proportional to the car’s width, specifically \(c = \frac{\text{car-width}}{4}\), where \(\pm 2\sigma\) covers 95\% of the Gaussian. \(m\) controls the DRF’s width during straight driving (\(\delta = 0\)), while \(k_1\) and \(k_2\) adjust the width in relation to the steering angle \(|\delta|\). This reflects the noise variability in human sensorimotor systems. Asymmetry in \(k_1\) and \(k_2\) accounts for different driving behaviors like curve-cutting, centerline following, or curve overshooting, improving risk management in tight curves. 

Thus, the DRF is parameterized by \(p\), \(t_{\text{la}}\), \(m\), \(c\), \(k_1\), and \(k_2\), and is solely dependent on the driver's state. Each object in the environment is assigned a cost, creating a cost map. This map is combined with the DRF by element-wise multiplication, summed across the grid to compute the QPR:
\begin{equation}
\text{QPR} = \sum_{\text{grid points}} (\text{Cost} \times \text{DRF})
\end{equation}
This metric reflects the driver's perceived likelihood and severity of potential incidents, bridging subjective perception with objective risk quantification.


\textbf{Omnidirectional Risk Quantification.} Traditional Driver Risk Fields (DRF) focus only on the forward-facing half-circle. To enable realistic risk assessment for autonomous driving, our model extends this to a 360-degree view, incorporating risks from both front and rear vehicles. By including the rear vehicle's DRF and its collision cost with the ego vehicle, we create a unified risk landscape from all angles, enhancing situational awareness and safety.

\begin{equation}
QPR_{\text{total}} = QPR_{\text{front}} + QPR_{\text{rear}}
\end{equation}

\begin{equation}
QPR_{\text{front}} = \sum_{\text{grid points}} (\text{Cost}_{\text{cars\_infront}} \times \text{DRF}_{\text{ego\_car}})
\end{equation}

\begin{equation}
QPR_{\text{rear}} = \sum_{\text{grid points}} (\text{Cost}_{\text{ego\_car}} \times \text{DRF}_{\text{cars\_behind}})
\end{equation}

Our approach not only calculates the total risk posed, but also evaluates the specific risk attributes of each participant. This enables the identification of those posing the greater danger, allowing for more targeted risk identification and warnings, as shown in \autoref{fig_qpr_attributes}.

\begin{algorithm}[h]
    \caption{Autonomous Driving by LLM-based Decision Agent with Risk Assessor}
    \label{alg:LLM}
    \KwInput{Scene description dataset \textit{scenes}, real-world next frame decision \textit{scene.true\_label}, risk assessment module \textit{RA}, reasoning module \textit{RM}, reflection module \textit{RF}, memory database \textit{memory}, few-shot number \textit{n}}
    
    \ForEach{\textit{scene} in \textit{scenes}}{
        {\fontsize{9}{10.8}\selectfont\tcp{Initialize scene description}}
        \textit{scene} = \textit{scene}
        
        {\fontsize{9}{10.8}\selectfont\tcp{Perform risk assessment}}
        \textit{risk\_info} = \textit{RA.assess\_risk(scene)} \\
        \textit{scene.add(risk\_info)}
        
        {\fontsize{9}{10.8}\selectfont\tcp{Retrieve few-shot examples}}
        \textit{few\_shots\_experience} = \textit{memory.retrieve(scene, few\_shots\_number = n)} \\     \textit{scene.add(few\_shots\_experience)}
        
        {\fontsize{9}{10.8}\selectfont\tcp{Perform reasoning to determine action}}
        \textit{reasoning\_process, action} = \textit{RM.reasoning(llm, scene)}
        
        {\fontsize{9}{10.8}\selectfont\tcp{Evaluate action and update memory}}
        \eIf{\textit{action} == \textit{scene.true\_label}}{
            {\fontsize{9}{10.8}\selectfont\tcp{Add GPT-4 outputs to the scene}}
            \textit{scene.add(reasoning\_process, action)}
            
            {\fontsize{9}{10.8}\selectfont\tcp{Update memory database}}
            \textit{memory.update(scene)}
        }{
            {\fontsize{9}{10.8}\selectfont\tcp{Perform reflection process}}
            \textit{reflection\_process} = \textit{RF.reflection(scene)}
            
            {\fontsize{9}{10.8}\selectfont\tcp{Update memory database}}
            \textit{memory.update(reasoning\_process, reflection\_process, action)}
        }
    }
    \KwOutput{Updated memory database}
\end{algorithm}

\begin{figure*}[ht]
    \centering
    \includegraphics[width=\textwidth]{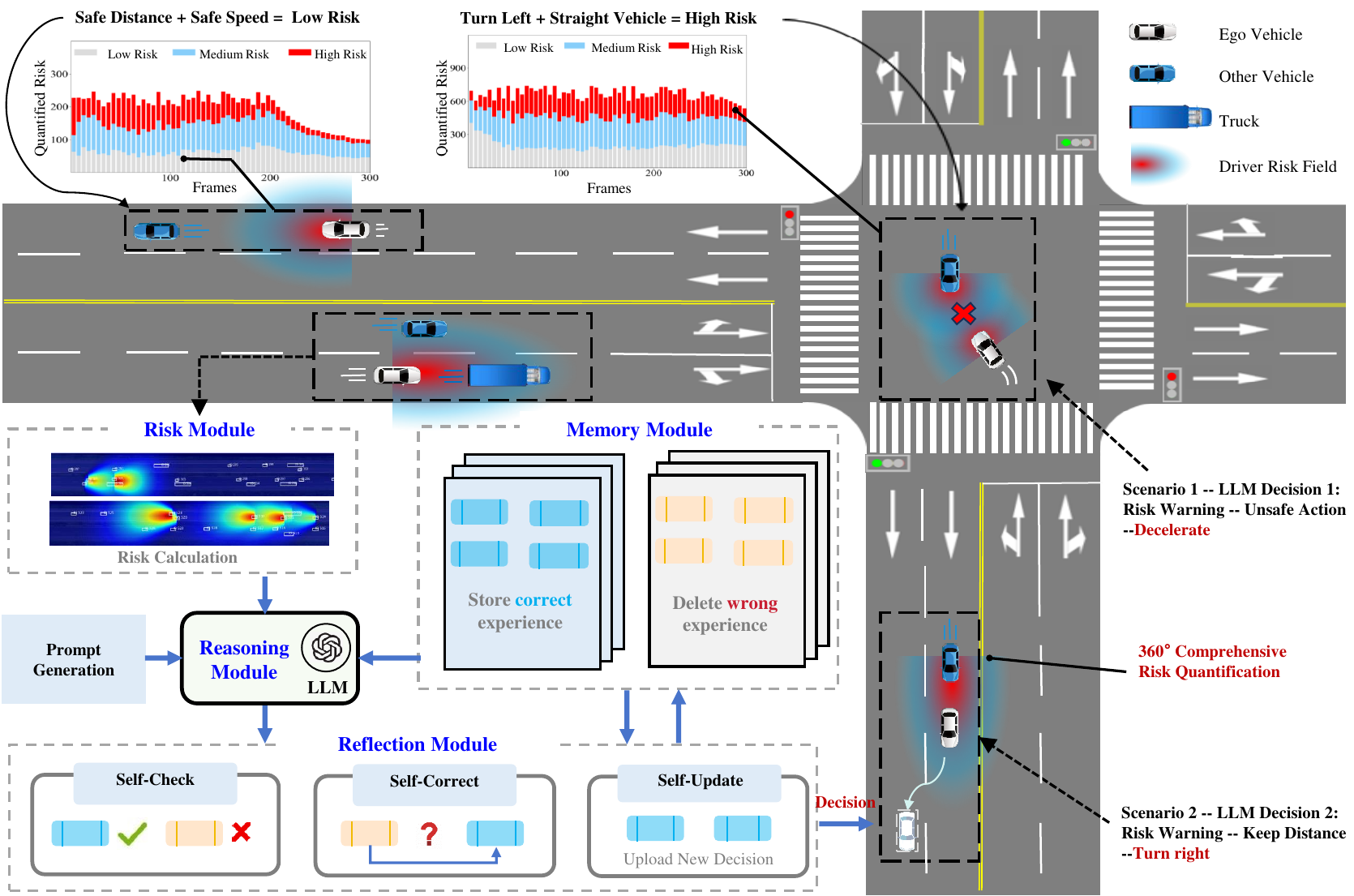}
    \caption{SafeDrive in real-world scenarios: The multi-dimensional risk quantification model captures comprehensive risk information from the surroundings. By combining this risk model with the closed-loop LLM-based driving agent, the system ensures both immediate risk prevention and implements long-term safety measures, ensuring real-time decision-making safety across diverse scenarios. The distributions of QPRs on highway and urban intersections are presented at the top left corner, and corresponding risk-levels of the driving scenes showcase the effectiveness and consistency of our model.}
    \label{fig_3}
\end{figure*}

\subsection{Risk-Sensitive LLM Decision-Making}
\textbf{RQ2. How can we guide the LLM-based agents to derive safe and human-like driving behavior?}

Building on the previously introduced risk quantification and prior knowledge-driven paradigm for autonomous systems, we leverage the reasoning power of large models to propose SafeDrive, a knowledge- and data-driven framework, as shown in \autoref{fig_3}. In this paper, GPT-4 serves as the decision-making agent, driving the reasoning process and generating actions. We use manually annotated scene descriptions, paired with next-frame actions as ground truth labels, from real-world datasets, including HighD (highways), InD (urban intersections), and RounD (roundabouts). These descriptions provide environmental context, such as the ID, position, and velocity of surrounding vehicles, enabling GPT-4 to interpret the environment and support reasoning and decision-making.

The SafeDrive architecture comprises four core modules: \textit{Risk Module}, \textit{Reasoning Module}, \textit{Memory Module}, and \textit{Reflection Module}. The process is iterative: the \textit{Reasoning Module} makes decisions based on system messages, scene descriptions, risk assessments, and store similar memories; the \textit{Reflection Module} evaluates decisions and provides a self-reflection process; and the \textit{Memory Module} stores correct decisions for future retrieval. Using three real-world datasets as input, this self-learning loop enhances decision accuracy and adaptability in handling of diverse, complex scenarios. The overall LLM-based decision algorithm is presented in \autoref{alg:LLM}.

\begin{figure*}[h]
    \centering
    \includegraphics[width=0.95\textwidth]{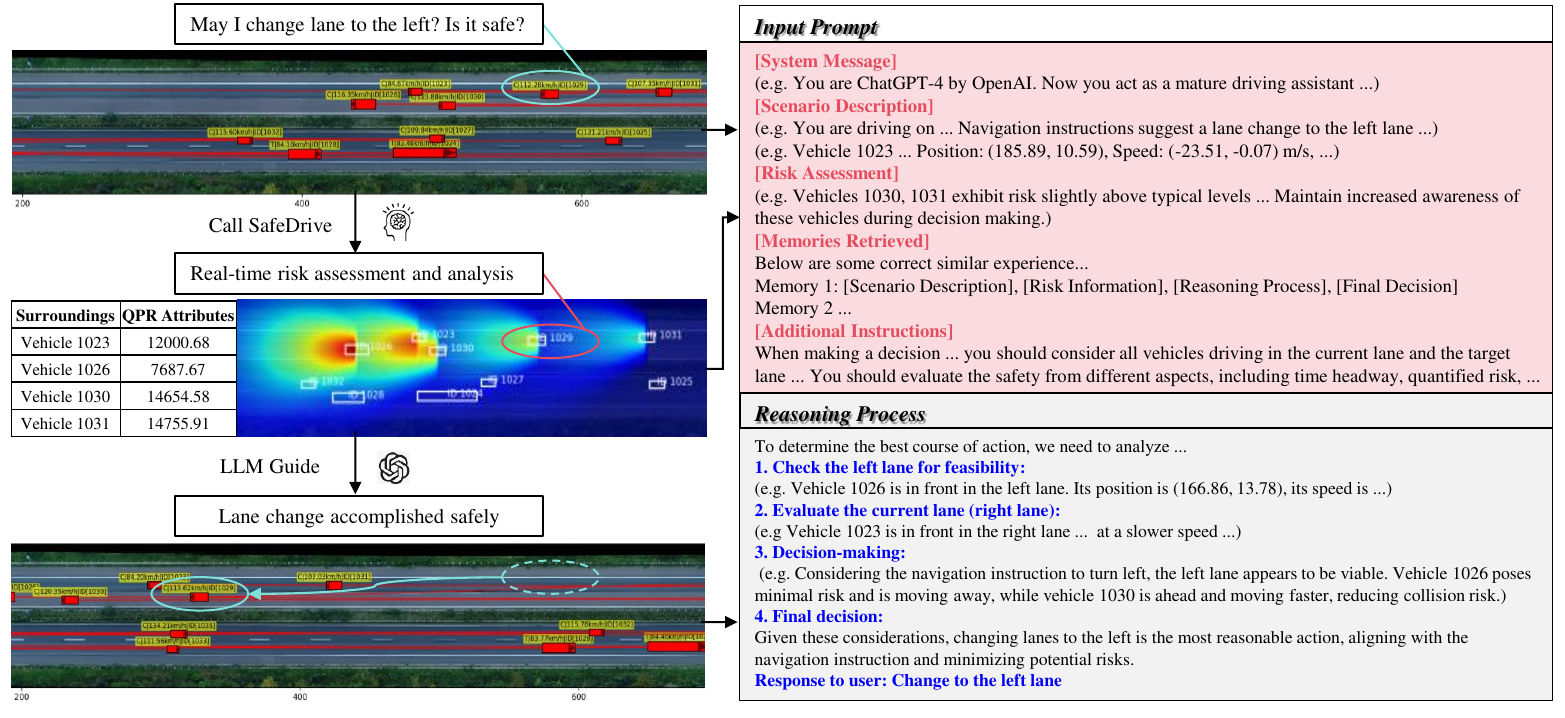} 
    \caption{Example of system prompts and interaction within the SafeDrive system.}
    \label{prompts}
\end{figure*}

As shown in \autoref{prompts}, in dynamic scenarios, SafeDrive receives user navigation instructions and scene descriptions, evaluating surrounding vehicles' risk attributes (e.g., QPR values), positions, and speeds in real time. Then leveraging LLM reasoning and historical memory, the system conducts feasibility checks, lane evaluations, and decision-making to determine the safest action, such as lane changes. Overall, by combining multi-dimensional risk quantification with GPT-4's reasoning, SafeDrive provides real-time, risk-sensitive decisions. In high-risk scenarios like highways and intersections, it identifies unsafe behaviors and makes adaptive decisions (e.g., deceleration or turning). A closed-loop reflection mechanism ensures continuous optimization, enhancing responsiveness, adaptability, and safety.

\textbf{Risk Module.}
The Risk Module generates detailed textual risk assessments for each participant based on the aforementioned risk quantification model and defined thresholds. These thresholds are experimentally determined, taking into account the risk distribution and common safety standards, addressing both longitudinal and lateral risks. This integration ensures heightened caution in decision-making, guiding the GPT-4 driving agent to effectively avoid or mitigate unsafe actions.

\textbf{Reasoning Module.} The reasoning module facilitates the system decision-making process through three key components. It begins with a system message that defines the GPT-4 driving agent’s role, outlines the expected response format, and emphasizes safety principles for decision-making. Upon receiving input composed of the scene description and risk assessment, the module interacts with the memory module to retrieve similar successful past samples and their correct reasoning processes. Finally, the action decoder translates output the decisions into specific actions for the ego vehicle, such as accelerating, decelerating, turning, changing lanes, or remaining idle. This structured approach ensures informed and safety-conscious decisions.

\textbf{Memory Module.} 
The memory module is a core component of our system, enhancing decision-making by leveraging past driving experiences. It stores vectorized scenarios using GPT embeddings in a vector database. The database is initialized with a set of manually created exemplars, each comprising a scene description, risk assessment, template reasoning process, and correct action. When encountering new scenarios, the system retrieves relevant experiences by matching vectorized descriptions using a similarity score. After the decision-making process, the new sample is added to the database. This dynamic framework enables continuous learning, allowing the system to adapt to diverse driving conditions.

\textbf{Reflection Module.} 
The reflection module evaluates and corrects erroneous decisions made by the driving agent, initiating a thinking process of why the agent chooses the wrong action. Corrected decisions and their reasoning are then stored in the memory module as references to prevent similar errors in the future. This module not only allows the system to evolve continuously but also provides detailed log information to developers, enabling them to analyze and refine the system message to improve the decision logic of the agent.

\begin{figure*}[h]
    \centering
    \includegraphics[width=0.95\textwidth]{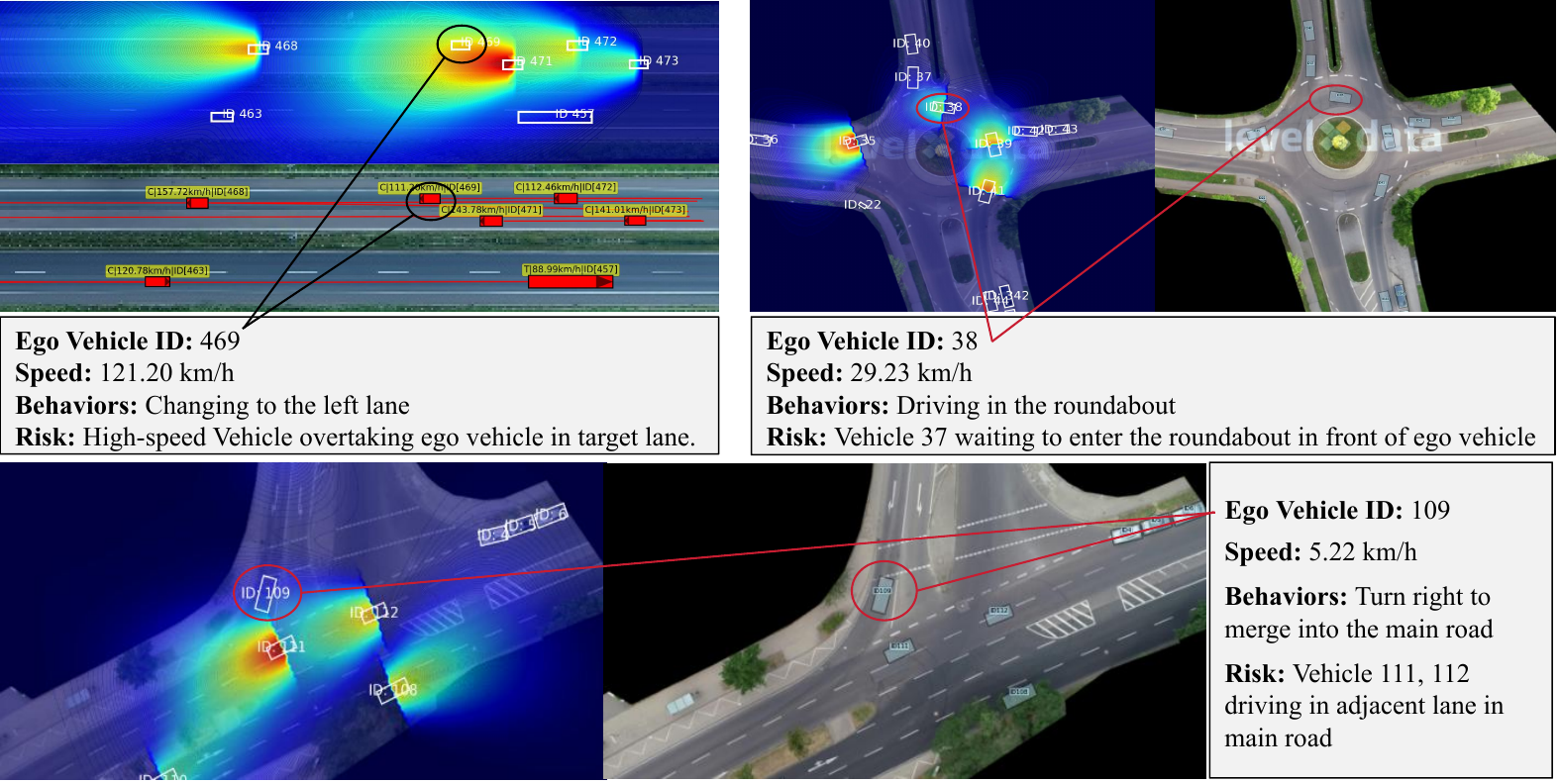} 
    \caption{Risk quantification heatmap in diverse driving scenarios: HighD, RounD, InD.}
    \label{multi-risk}
\end{figure*}

\section{Experiment Results}

The experiment section of this study is divided into two parts: one focusing on the robustness of the risk quantification model and the other on the performance of SafeDrive.

\subsection{Performance on Risk Quantification}

\textbf{Datasets Preparation.} To evaluate the robust performance of our model across diverse environments, we used the HighD~\cite{highd}, RounD~\cite{round}, and InD~\cite{ind} datasets. The HighD dataset (60 recordings from six highways) tested the model’s handling of car-following and lane-changing on highways. The InD dataset (11,500 trajectories at urban intersections) assessed its ability to manage diverse participants and unpredictable urban behaviors. The RounD dataset (13,700 road users in roundabouts) evaluated its performance in high-risk maneuvers and interactions. Together, these datasets provided a robust framework for testing the model across highways, intersections, and roundabouts, ensuring reliability in varied real-world scenarios.

\begin{figure*}[h]
    \centering
    \includegraphics[width=0.95\textwidth]{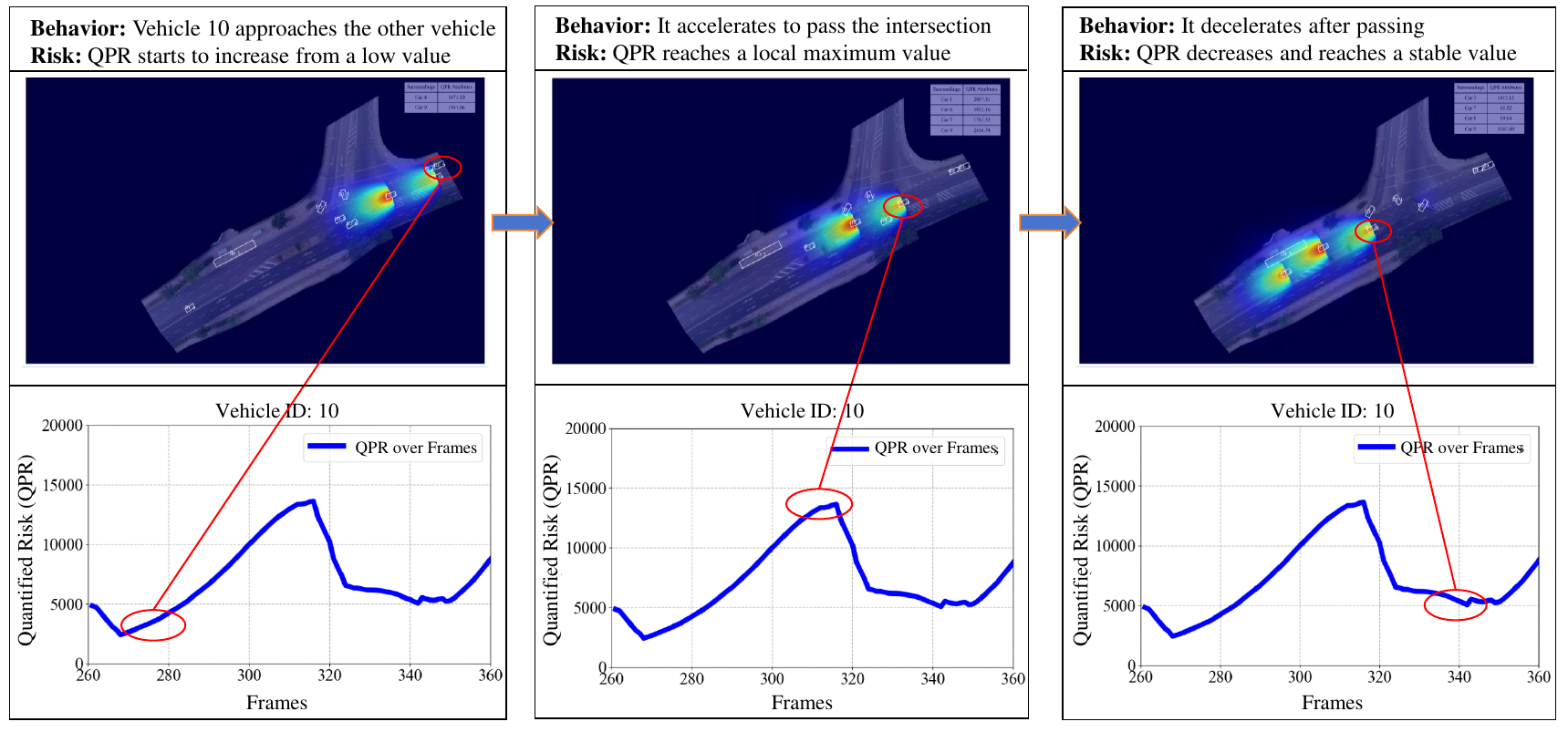} 
    \caption{The QPR variation of vehicle passing an urban intersection.}
    \label{risk-change}
\end{figure*}

\textbf{Risk Quantification Process.} The process is designed to accurately capture and assess the intricate risk behind real-world driving scenarios, as shown in \autoref{multi-risk}. It presents autonomous driving risk quantification and behavioral analysis across three distinct driving scenarios: a highway, a roundabout, and an intersection. For instance, in the highway scenario, the ego vehicle (ID 469) performs a left lane change while facing the risk of high-speed overtaking vehicles in the target lane. These scenarios are visualized with risk quantification heatmaps that clearly highlight high-risk areas, providing critical support for decision-making optimization and safety enhancement.

First, we demonstrate that our risk quantification aligns with common traffic logic. The continuous QPR variation of the ego vehicle as it passes through an urban intersection in the InD dataset. The QPR increases as the vehicle approaches the intersection with other vehicles waiting to merge, and returns to normal levels after the ego vehicle clears the intersection. This behavior accords with common traffic sense. Furthermore, if we integrate this result into a control system, the ego vehicle will decelerate as it approaches the intersection to mitigate the risk, then idle or accelerate after passing when the QPR normalizes. This mirrors a safe, human-like driving behavior, which is a desirable property for autonomous driving system, as shown in \autoref{risk-change}.

Second, we analyzed the QPR distribution using 500,000 random samples from surrounding vehicles. Thresholds for low, medium, and high risk levels were determined based on the 30th and 70th percentiles: risks below the 30th percentile were classified as low, between the 30th and 70th as medium, and above the 70th as high. It ensures accurate risk assessment and effective conversion into prompts for LLM input. 

\begin{figure*}[h]
    \centering
    \includegraphics[width=\textwidth]{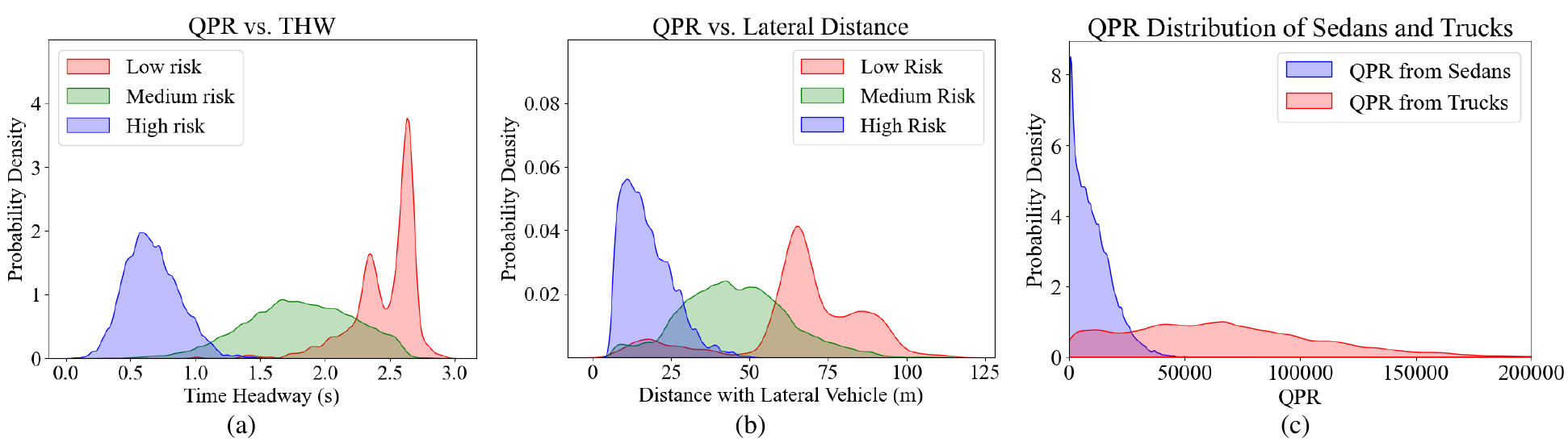} 
    \caption{The QPR analysis across THW, lateral distance, and vehicle types.}
    \label{QPR vs. THW}
\end{figure*}

To validate the reliability of our established thresholds, we compared them with commonly accepted safety metrics. We analyzed QPR against THW to evaluate longitudinal risk quantification. A safe THW, typically considered to be 2 seconds, helps maintain adequate following distances and reduces the risk of potential collisions. As shown in \autoref{QPR vs. THW} (a), most samples categorized as low risk fall within this range. Furthermore, the distinct separation between the three risk levels in the graph demonstrates the effectiveness of our model in quantifying longitudinal risk.

In complex multi-lane environments, vehicles in adjacent lanes or at intersections significantly affect driving decisions and safety. Unlike traditional systems focused on longitudinal safety, our model incorporates lateral risk quantification. As shown in \autoref{QPR vs. THW} (b), the relationship between QPR and lateral distance confirms the model's effectiveness. By accounting for both longitudinal and lateral risks, the system predicts and mitigates side maneuver and lane-change risks, frequent causes of accidents. Our risk model differentiates between various road participants, such as sedans and trucks, to account for their unique risk profiles. Trucks, for instance, are classified as higher risk due to their larger size and the more severe consequences of potential collisions. \autoref{QPR vs. THW} (c) demonstrates that our model effectively captures these nuances, enhancing the precision and safety of its responses in complex urban environments.

\begin{table}[h!]
\centering
\caption{Details of Experiments Set-ups}
\begin{tabular}{lccc}
\toprule
\textbf{\centering Scenarios} & \textbf{\centering Response Time} & \textbf{\centering Tokens per Episode} & \textbf{\centering \# Memory} \\
\midrule
HighD  & 20.68s ± 3.68s & \( \approx \)330,000 & 3 \\
InD    & 18.63s ± 3.07s & \( \approx \)340,000 & 3 \\
Round  & 24.12s ± 3.52s & \( \approx \)280,000 & 2 \\
\bottomrule
\end{tabular}
\label{tab:experiment_setup}
\end{table}

\subsection{Performance on Reasoning and Decision-Making}

\begin{figure*}[h]
\centerline{\includegraphics[width=1\textwidth]{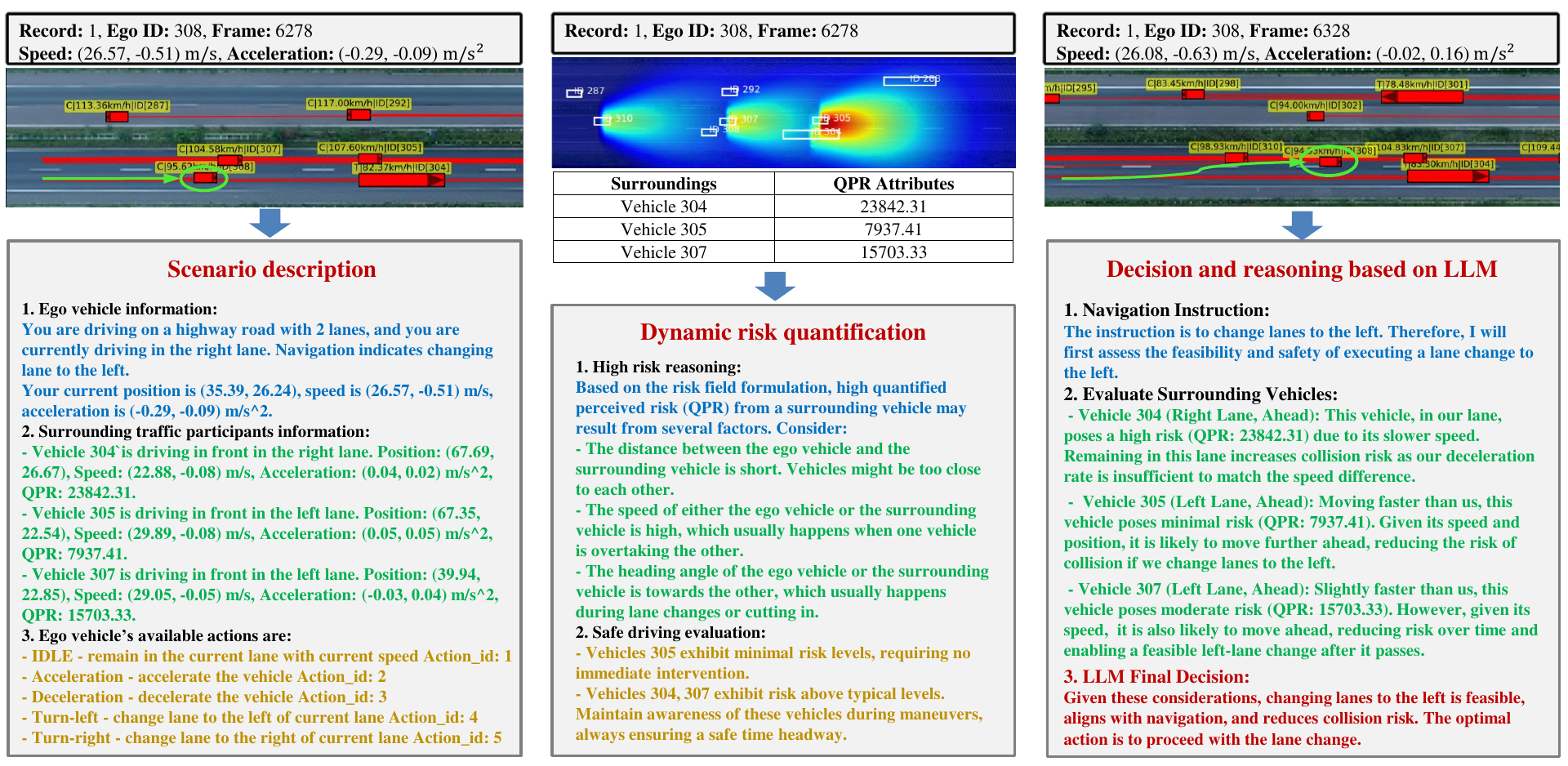}}
		\caption{Visualization of a basic decision-making process (no memory \& reflection). Scenario description \textbf{(Left)}: Id, coordinate (and lane position), and dynamic information of ego vehicle and other vehicles. Risk assessment \textbf{(Mid):} QPR and detailed textual risk notification regarding each surrounding vehicle. Decision-making \textbf{(Right):} GPT-4 reasoning with the prior two parts as input.}
		\label{LLM reasoning}
	\end{figure*}

We begin by showcasing a sample of the decision-making process. As shown in \autoref{LLM reasoning}, our agent successfully follows navigation instructions while maintaining safety. The process starts with scenario description, which outlines the vehicle's state, traffic dynamics, and available actions. Dynamic risk quantification follows, using heatmaps and textual notifications to identify high-risk traffic components. Finally, the reasoning module integrates these inputs to assess lane-change feasibility and determine the optimal driving strategies. By incorporating all relevant environmental details, this streamlined process ensures comprehensive decision-making.

\begin{figure*}[h]
\centerline{\includegraphics[width=1\textwidth]{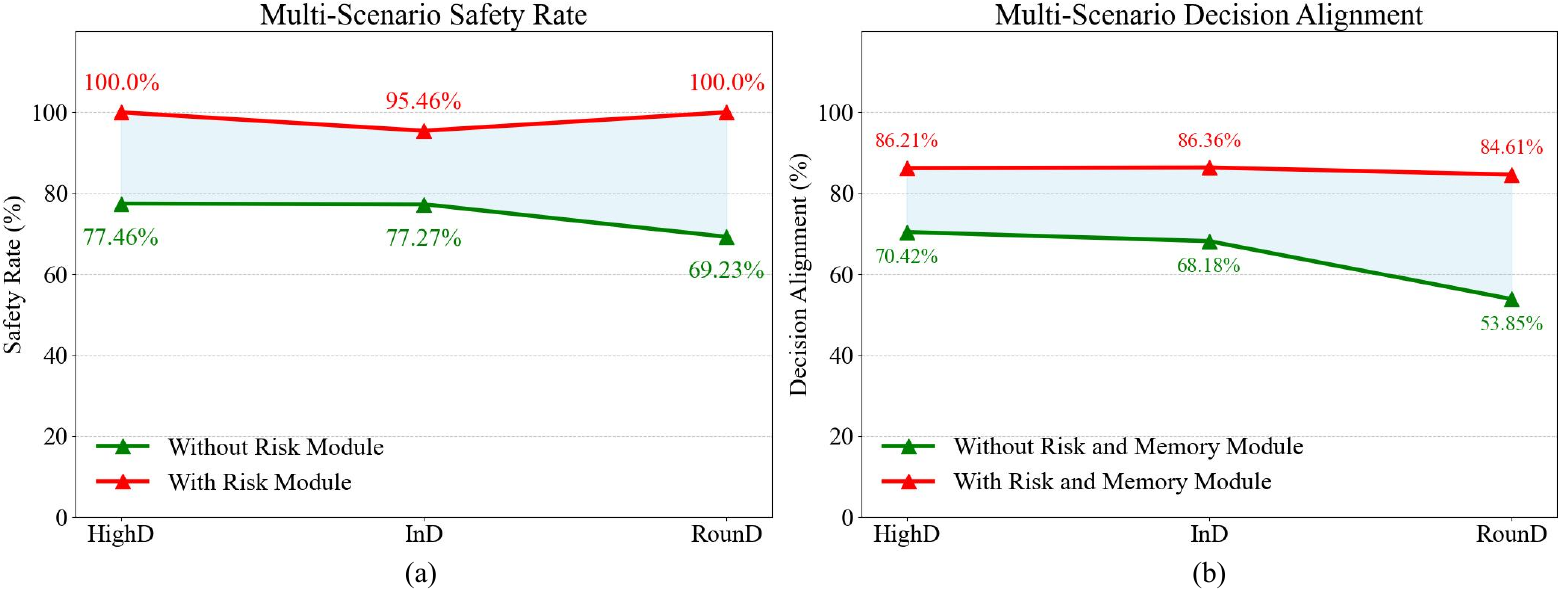}}
		\caption{Safety rate \textbf{(left)} and decision alignment percentage \textbf{(right)} of three scenarios: substantial improvements in both measures.}
		\label{experiment improvement}
	\end{figure*}

\begin{figure}[h] 
\centering 
\includegraphics[width=\columnwidth]{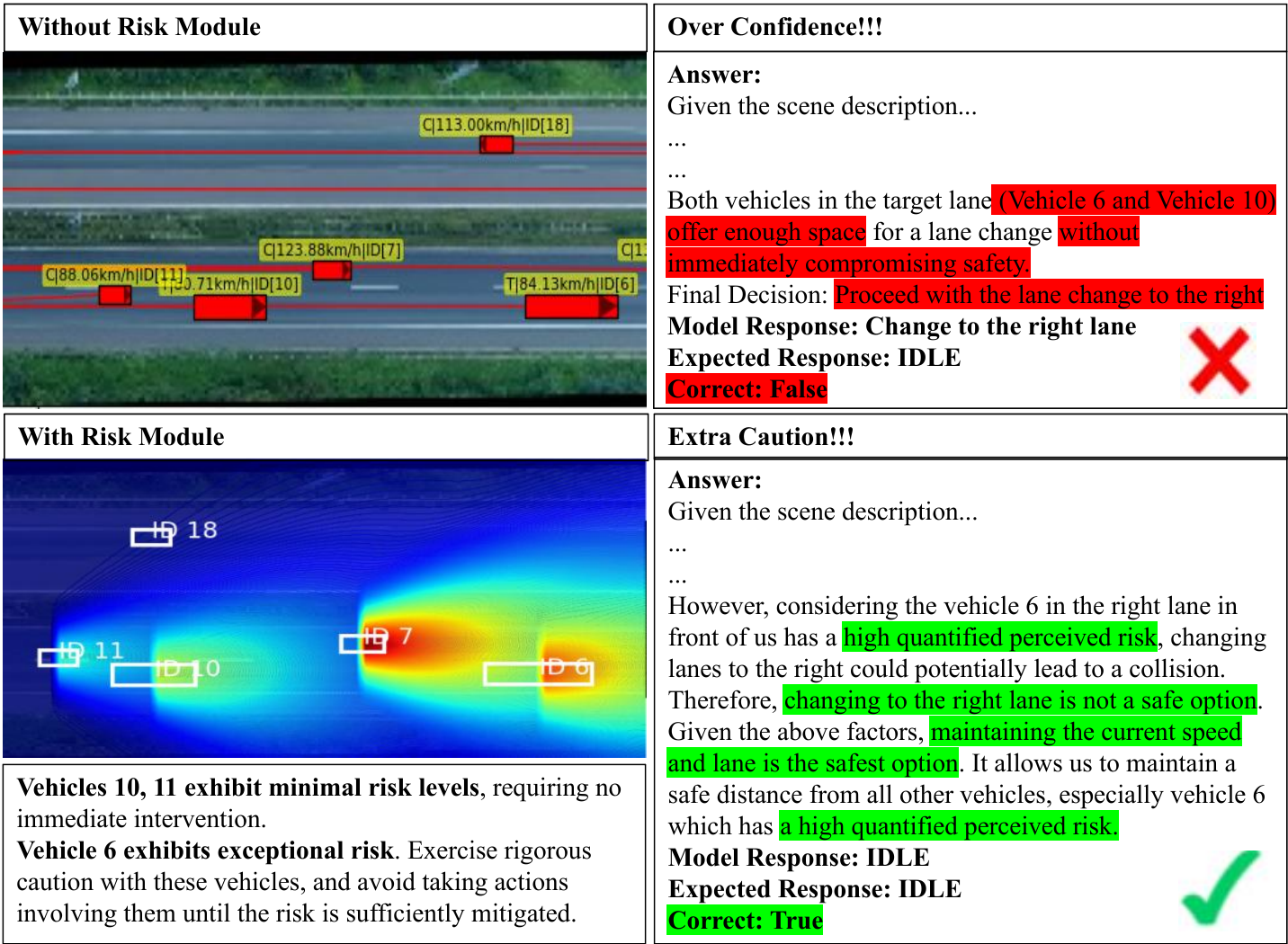} 
\caption{The risk module prevents the generation of dangerous actions. \textbf{Top:} Decision-making with the risk module. \textbf{Bottom:} Decision-making without the risk module.} 
\label{with_safety} 
\end{figure}

Our results are based on the average performance over ten experimental episodes. \autoref{tab:experiment_setup} presents the average response time, token count, and number of pre-loaded memories.

\textbf{Safety Rate Improvement}: 
The safety rate  is defined as the probability that a vehicle takes safety measures to avoid accidents.
We conduct a comparative analysis of our method against several standard approaches and evaluate the system's safety performance with and without the Risk Module using the same datasets. An action is deemed safe if it does not disrupt the behavior of other vehicles, such as causing sudden braking or unexpected maneuvers. Specifically, IDM \cite{kesting2010enhanced} employs the Intelligent Driver Model to maintain safe distances from other agents and adhere to the reference lane.  

As shown in \autoref{experiment improvement} (a), and  \autoref{tab:IntegratedMetrics}, for Safety Rates, IDM achieves 76.00\%, 92.50\%, and 86.67\% on the InD, HighD, and RounD datasets, respectively, while GPT-4 performs lower at 77.27\%, 77.46\%, and 69.23\%. Adding the \textit{Memory Module} improves safety rates to 86.36\%, 81.84\%, and 80.77\%. With the \textit{Risk Module}, safety rates reach 100\% in HighD and RounD, and 95.46\% in InD. Combining both modules achieves 100\% safety across all datasets, demonstrating comprehensive safety assurance in diverse and complex scenarios.

These enhancements highlight the \textit{Risk Module}'s effectiveness in improving system safety, with no unsafe actions generated throughout the entire experimental process. The results further emphasize the module's adaptability and risk mitigation capabilities, ensuring robust performance across different driving environments, even in challenging conditions.

\autoref{with_safety} presents a typical example demonstrating the safety enhancement of our framework. It compares performance with and without the safety module. Without the safety module, the model shows overconfidence in high-risk scenarios, failing to identify potential threats, leading to unsafe lane-change decisions. In contrast, with the safety module, the GPT-4 agent better understands potential risk sources in the environment, accurately identifying high-risk targets (e.g., Vehicle 6) and making more cautious decisions (e.g., staying in the current lane). This extra caution provided by risk module helps prevent the agent generate risky behaviors, thus significantly improving decision safety and reliability.

\begin{table*}[htbp]
\centering
\caption{Experimental Results for InD, HighD, and RounD Scenarios}
\begin{threeparttable}
\begin{tabular}{lcccccc}
\toprule
\textbf{Scenario} & \textbf{Metrics} & \textbf{IDM \cite{kesting2010enhanced}} & \textbf{GPT-4} & \textbf{With Memory Module} & \textbf{With Risk Module} & \textbf{With Both Modules} \\
\midrule
\multirow{2}{*}{InD}   & Safety Rate  & 71.21\%     & 77.27\% & 86.36\% & 95.46\% & 100\% \\
                       & Decision Alignment & 60.61\% & 68.18\% & 81.82\% & 63.29\% & 86.36\% \\
\midrule
\multirow{2}{*}{HighD} & Safety Rate   & 91.36\%     & 77.46\% & 81.84\% & 100\%   & 100\% \\
                       & Decision Alignment & 75.31\% & 70.42\% & 87.32\% & 61.41\% & 86.21\% \\
\midrule
\multirow{2}{*}{Round} & Safety Rate  & 84.62\%     & 69.23\% & 80.77\% & 100\%   & 100\% \\
                       & Decision Alignment & 69.23\% & 53.85\% & 76.93\% & 69.23\% & 84.61\% \\
\bottomrule
\end{tabular}
\end{threeparttable}
\label{tab:IntegratedMetrics}
\end{table*}

\textbf{True Decision Alignment Enhancement:} 
The decision alignment refers to the percentage of decisions output by the LLM agent that align with the subsequent real-world human driver decisions.
In this part of the experiment, we conducted tests to compare the system's decision accuracy with and without both the risk and memory modules, using the same datasets for both conditions. A generated decision is considered correct if it aligns with the true actions taken by real-world human drivers in the subsequent few frames in our hand-labeled datasets.

By integrating the Risk Module and Memory Module, the system's decision alignment percentage has increased significantly, as demonstrated in \autoref{experiment improvement} (b) and  \autoref{tab:IntegratedMetrics}. For decision alignment, IDM achieves 72.00\%, 76.25\%, and 73.33\% on the InD, HighD, and RounD datasets, respectively. GPT-4 decreases to 68.18\%, 70.42\%, and 53.85\%. Adding the \textit{Memory Module} significantly improves alignment to 81.82\%, 87.32\%, and 76.93\%. While the \textit{Risk Module} alone performs slightly lower, combining both modules achieves 86.36\%, 86.21\%, and 84.61\%, demonstrating the best consistency and stability.
This improvement highlights the effectiveness of our framework in achieving decisions aligned with real-world optimal outcomes. By integrating risk assessment and memory recall, the system mimics human-like decision-making, ensuring safer and more reliable performance. The high alignment across diverse datasets—highway, urban intersections, and roundabouts—demonstrates the framework's robustness and adaptability in navigating dynamic traffic environments and varying decision-making challenges.

During our experiments, we observed that adjusting the system's focus on risk levels through modified prompt occasionally led the agent to exhibit higher-level human-like driving behavior. For instance, in a scenario where a following vehicle overtakes the ego vehicle at high speed on a highway, increasing the LLM's focus on quantified risk level guided the system to shift to the right lane, clearing the path for the following vehicle and thus mitigating the risk. This response mirrors the behavior of experienced human drivers on highways, further demonstrating the effectiveness of our risk quantification model in emulating such actions. This surprising finding indicates that our system not only addresses immediate risks, but also fosters decisions that consider long-term safety, a valuable property for AVs. Such higher-level, human-like behavior is a critical feature, positioning our system as a versatile solution for advancing autonomous driving.

\section{Conclusion}
In this paper, we introduce SafeDrive, a knowledge- and data-driven framework for AVs' risk-sensitive decision-making, addressing the critical challenges of safety in unpredictable, high-risk and long-tail scenarios. By proposing a unified risk quantification model capable of omnidirectional evaluation of multi-factor coupled risks, the framework effectively models the complexities of stochastic urban environments (RQ1). Integrating this risk model into an LLM-driven decision-making framework, powered by GPT-4, enables safe, human-like decision-making and continuous adaptability in uncertain traffic conditions (RQ2). Extensive evaluations using real-world datasets (HighD, InD, and RounD) demonstrate the framework's robust performance, achieving a 100\% safety rate and over 85\% alignment with human driving behaviors. This highlights the framework's effectiveness in replicating human-like strategies while ensuring long-term reliability across diverse and challenging scenarios.

In the future, we will enhance the risk quantification model by incorporating more detailed environmental features, such as road boundaries and obstacles. Additionally, transitioning to advanced LLMs with improved reasoning capabilities, such as ChatGPT o1-preview \cite{wu2024comparative}, will enable more nuanced decision-making. By leveraging fine-tuning techniques emphasized in recent research \cite{Sun_2024}, we aim to further improve the system's expertise in specific domains, such as traffic knowledge, making it a more mature and capable driving agent.

\section*{Acknowledgment}

We thank the authors of the paper "DiLu: A Knowledge-Driven Approach to Autonomous Driving with Large Language Models" (Licheng Wen et al.) for their valuable work and code, which greatly contributed to this research.


\bibliographystyle{IEEEtran}
\bibliography{main}\ 

\end{document}